\documentclass[runningheads]{llncs}

\usepackage{eccv}

\usepackage{eccvabbrv}

\usepackage{graphicx}
\usepackage{booktabs}

\usepackage{amsfonts}
\usepackage{comment}
\usepackage{bbm}

\usepackage[fixed]{fontawesome5} %
\newcommand{\Cow}[1][]{\includegraphics[width=10pt,trim={4cm 6cm 3cm 3cm},clip]{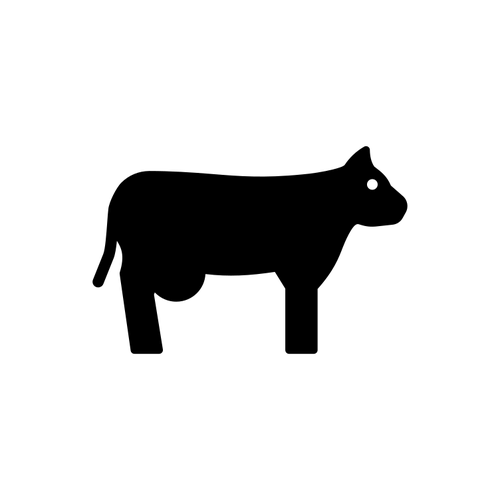}}
\newcommand{\Plant}[1][]{\includegraphics[width=10pt,trim={5cm 4cm 3cm 1cm},clip]{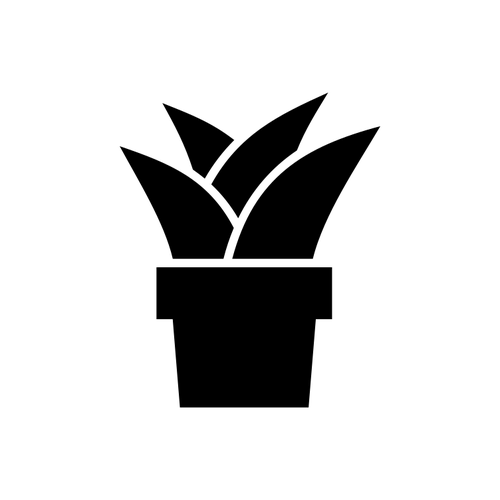}}
\newcommand{\Sheep}[1][]{\includegraphics[width=10pt,trim={4cm 4cm 3cm 3cm},clip]{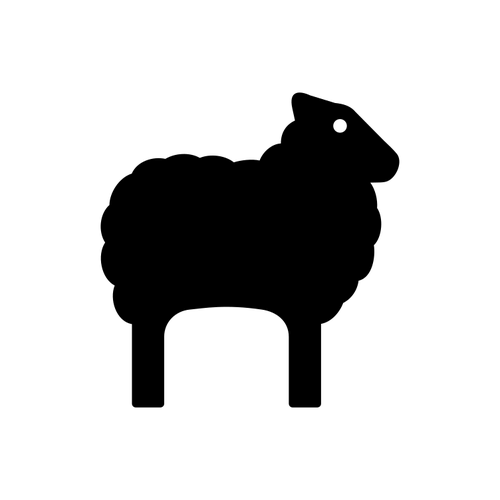}}

\usepackage[accsupp]{axessibility}  %

\usepackage{hyperref}

\usepackage{orcidlink}

\begin{document}

\title{Self-Supervised Learning of Semantically Consistent Visual Features via a Simple 3D Prior}

\titlerunning{Self-Supervised Semantically Consistent Features via a Simple 3D Prior}

\author{Sebastian Day~\orcidlink{0009-0003-5119-3957} \and
Oisin Mac Aodha~\orcidlink{0000-0002-5787-5073}}

\authorrunning{S. Day et al.}

\institute{University of Edinburgh}

\maketitle

\begin{abstract}
Visual features extracted from vision models like DINOv2 or Stable Diffusion enable high quality semantic correspondence between images. 
However, such features exhibit documented limitations that hinder accurate correspondence estimation such as the inability to disambiguate the left from right for categories exhibiting symmetries or repeated parts. 
Previous attempts to improve these features rely on using supervised data such as annotated keypoint pairs or camera pose information, or make restrictive topological assumptions, such as modelling the object of interest as a 3D sphere. 
We introduce a new self-supervised approach that allows us to learn features for topologically complex object categories using a simple prior. 
Instead of relying on an expensive instance-level analysis by synthesis pipeline, our approach makes use of contrastive learning and distribution matching at the global dataset-level to learn the coarse  shape and appearance of a category. 
Combined with additional geometric losses, this enables us to replicate the global and local statistics of visual features which match those that are extracted from real images. 
Quantitative results are presented across several challenging object categories, demonstrating the effectiveness of our approach. 
\end{abstract}

\vspace{-5pt}
\section{Introduction}
\vspace{-5pt}
\label{sec:intro}

Being able to decompose complex articulated and deformable object categories into structured representations that encode semantics is a key desirable capability for embodied AI systems. 
Semantic correspondence estimation is a specific instance of this task where the goal is to match the same local semantic parts/features across different instances of the same object category. 
The ability to find such correspondences is a key enabling technology for a variety of applications including robotic manipulation~\cite{tsagkas2024click}, image generation and editing~\cite{kim2024stableviton}, visual recognition~\cite{tang2020revisiting}, among others.

The previous most reliable recipe for this task involved training supervised learning-based approaches using large collections of pre-defined known semantic keypoint annotations, \eg the joints on the human body~\cite{guler2018densepose}. 
However, recently it has been demonstrated that general purpose features obtained from large vision models trained using self-supervised~\cite{caron_emerging_2021,oquab_dinov2_2023} or generative~\cite{rombach_high-resolution_2022} objectives are highly effective for semantic correspondence estimation, often surpassing features obtained from bespoke supervised methods~\cite{zhang_tale_2023}. 
These features exhibit several key advantages: they do not require category or semantic keypoint specific fine-tuning, correspondences can be efficiently computed using distance-based matching, and they generalise better than supervised methods~\cite{mariotti2025jamais}.

Despite the effectiveness of self-supervised features in semantic correspondence a prominent failure case is that of models becoming confused when there are repeated instances of a part, \eg confusing the front and back wheel of a car~\cite{zhang2024telling}.  
One hypothesis to explain this behaviour is that the representations learned by these large models fail to encode information about the underlying 3D shape or structure of objects~\cite{mariotti_improving_2024}. %
An obvious solution to this problem is to try to inject 3D knowledge into the features. 
Various techniques to attempt this have been proposed ranging from methods that use very simplistic, and thus inexpressive, geometric priors~\cite{mariotti_improving_2024}, all the way to methods that estimate category-specific 3D articulated meshes~\cite{wu_magicpony_2023}. 
These approaches typically require some form of additional supervision, whether it is approximate camera viewpoints~\cite{mariotti_improving_2024} or category-specific template meshes~\cite{shtedritski_learning_2023}. %

\begin{figure}[t]
    \centering
        \includegraphics[width=0.85\linewidth,trim={1.2cm 1.5cm 0.5cm 0.5cm},clip]{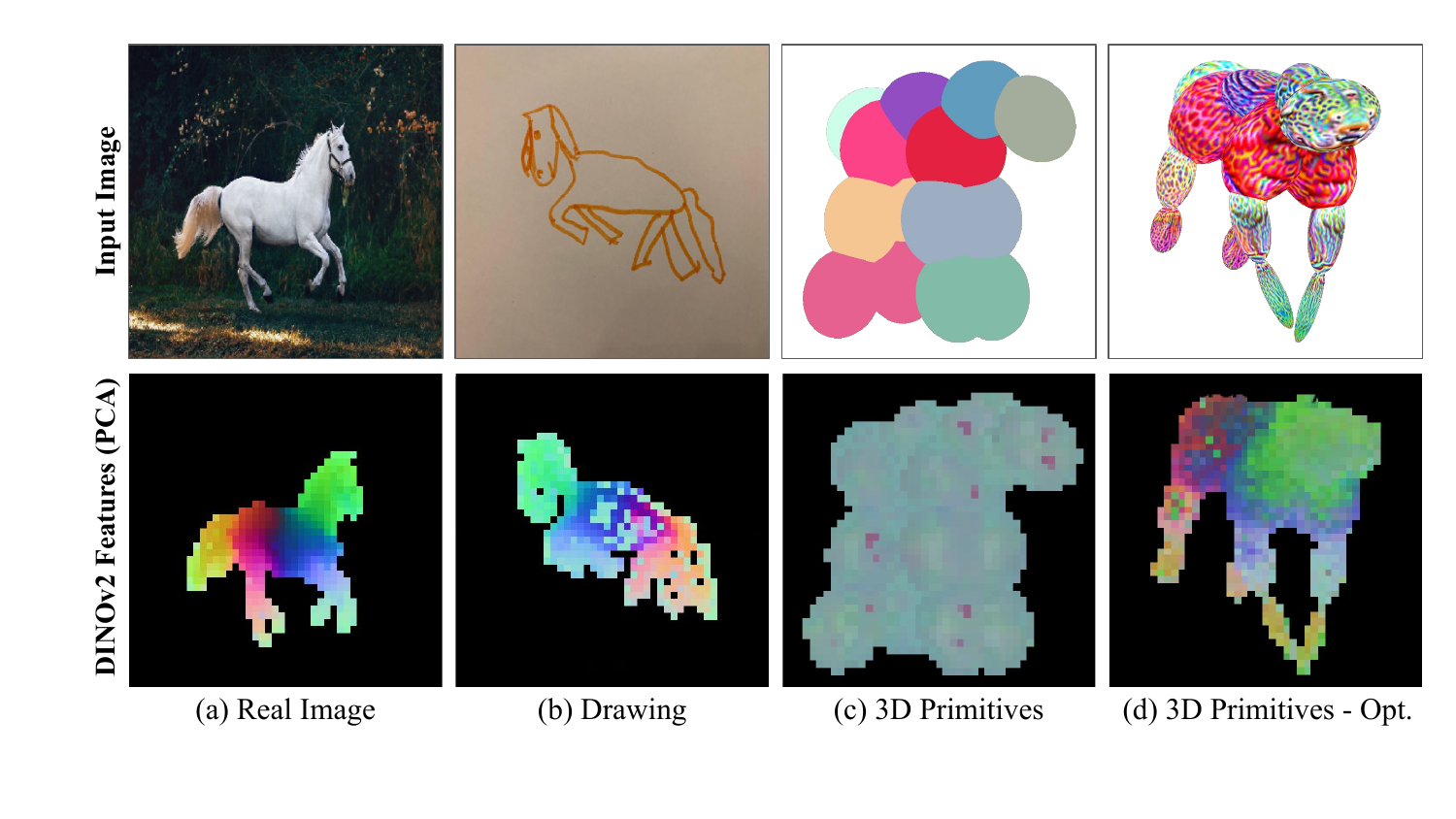}
        \vspace{-5pt}
        \caption{Pre-trained self-supervised models like DINOv2~\cite{oquab_dinov2_2023} produce local visual features that encode distinct semantic parts that generalise across different depictions of a category. 
        In (a) and (b) (images from~\cite{oquab_dinov2_2023}), we observe that the features for the horse's head and tail are similar even if when coming from an image or drawing. 
        In (c) we display the features derived from an abstract 3D collection of spheres that broadly has the same structure as a horse. 
        However, the resulting features are not consistent with those from (a) and (b). 
        In (d) we show features from our proposed self-supervised approach that optimises the shape and appearance of the shape from (c) such that when rendered and passed through DINOv2 they result in consistent visual features.     
        }
        \vspace{-15pt}
    \label{fig:motivation}
\end{figure}

We propose a method for extracting visual features that is more expressive than using a single simple 3D primitive, yet much more efficient to estimate than a full complex articulated 3D shape. 
Inspired by how well DINOv2 features~\cite{oquab_dinov2_2023} generalise across different visual depictions (see Fig.~\ref{fig:motivation}), our method leverages a simplistic synthetic 3D representation and optimises it to match the features extracted from real images.
Specifically, given an initial prior, our self-supervised approach constructs a simple 3D mesh that is represented as ellipsoids. 
This mesh and surface texture are learned such that when rendered to an arbitrary view and input to a feature extractor, it results in visual features that are similar to those obtained from real images. 
Rather than per-instance optimisation as in analysis-by-synthesis approaches, our approach is lightweight as we make use of geometric losses and minibatch-level feature distribution matching during training. 
We demonstrate that our approach works with either of two minibatch-level learning algorithms and, while only having relatively few learnable parameters, can generalise to a wide variety of categories (\eg quadrupeds,  spiders, bicycles, etc.), and is robust to variation across instances. 
We report quantitative results for semantic correspondence across different object categories  and demonstrate that our approach can learn semantically-aligned visual features.

\vspace{-5pt}
\section{Related Work}
\vspace{-5pt}

Most recent works on the task of semantic correspondence estimation usually follow a similar approach of taking a pre-trained feature detection model, such as the feature layers of DINOv2~\cite{oquab_dinov2_2023, simeoni_dinov3_2025} or Stable Diffusion~\cite{rombach_high-resolution_2022}, and train an adapter of the features. 
Features can be used directly or aggregated across multiple layers \cite{amir_deep_2022, zhang2024telling}. 
Features from vision-language models (VLMs) have also been used for semantic keypoint detection \cite{zhang_open-vocabulary_2024}. 

\noindent\textbf{Supervised methods.} 
These approaches typically perform contrastive learning on features extracted from annotated keypoints \cite{zhang2024telling, tang_emergent_2023, zhang_tale_2023, avidan_demystifying_2022,luodiffusion2023}. 
\cite{zhang2024telling} uses SoftArgMax on feature distances and an L2 penalty on distance from the target keypoint to post-process pre-trained features using contrastive learning. 
Pose graphs have also been explored in~\cite{hirschorn_graph-based_2025}. 
The limitation of supervised approaches is they require extensive expensive to obtain manual annotations. 
They also have a tendency to overfit to the trained keypoints and hence do not generalise~\cite{mariotti2025jamais}.

\noindent\textbf{Unsupervised methods.} 
This family of methods focus on taking features from pre-trained  models and adapting them through various means, \eg congealing \cite{ofri-amar_neural_2023, gupta_asic_2023}, contrastive learning~\cite{xiao2022learning}, or imposing structure \cite{barel_spacejam_2025, mariotti_improving_2024}.  
Techniques have been proposed that rely on self-consistency objectives, using either affine image transformations or known similar keypoints between images to push or pull features toward or away from one another \cite{avidan_demystifying_2022, zhang2024telling, karmali_lead_2022, ofri-amar_neural_2023, gupta_asic_2023, peebles_gan-supervised_2022}. 
Some works attempt to establish geometric transformations between images by applying image-based~\cite{shtedritski_learning_2023} or latent-based \cite{peebles_gan-supervised_2022} transformations and using the transformed points for self-supervised consistency. 
The transformations cannot allow too large a deviation from the original and hence the geometric gain can be limited. 

Some of these methods are weakly supervised as they use segmentation, pose labels or depth maps \cite{mariotti_improving_2024, mariotti2025jamais}.
Geometric constraints can be added such as flow and rigidity~\cite{barel_spacejam_2025}, and the learning can be constrained to a canonical mesh~\cite{leonardis_shic_2025, neverova_continuous_2020}.
 Multi-view images with pose labels can be used to learn 3D Gaussian representations which can then be used to supervise downstream tasks~\cite{yue_3daware_2024}. 
However, obtaining representative meshes for uncommon categories can be challenging. 
Canonical spaces can be more abstract, for the purposes of relating features to one another, or storing canonical pixel values, taking the form of a grid or sphere \cite{mariotti_improving_2024,gupta_asic_2023,ofri-amar_neural_2023}, or use point cloud representations requiring depth maps \cite{mariotti2025jamais, wandel_semalign3d_2025}.

Reconstruction-based methods can avail of large image collections and  use them to estimate a 3D model of a category \cite{aygun_saor_2024, wu_magicpony_2023, li_learning_2024} that can be optimised per image \cite{yao_lassie_2022, yao_hi-lassie_2023}. 
These methods both use pre-trained visual features from real images for training but can be used to obtain new features using synthetic renders of the learned shape~\cite{kaye_dualpm_2025}.
However, these models can be brittle and they require carefully constructed training pipelines and are limited to the categories they are trained on. 
They struggle with more complex shapes (\eg  spiders) when deforming the template mesh.
Diffusion models have been used to generate synthetic training images for these models~\cite{jakab_farm3d_2024}, but it is difficult to control the articulation and 3D pose, and the renders can hallucinate structures (\eg additional limbs). 
DualPM~\cite{kaye_dualpm_2025} can reconstruct accurate 3D point clouds for single images but it requires a per-category mesh with manually specified pose articulations.
Similar to \cite{shtedritski_learning_2023}, we also train a feature adapter, but our approach learns the explicit 3D model that enforces a plausible feature to 3D mapping.

\vspace{-5pt}
\section{Method}
\vspace{-5pt}

\begin{figure}[t]
\centering
\includegraphics[width=0.92\textwidth]{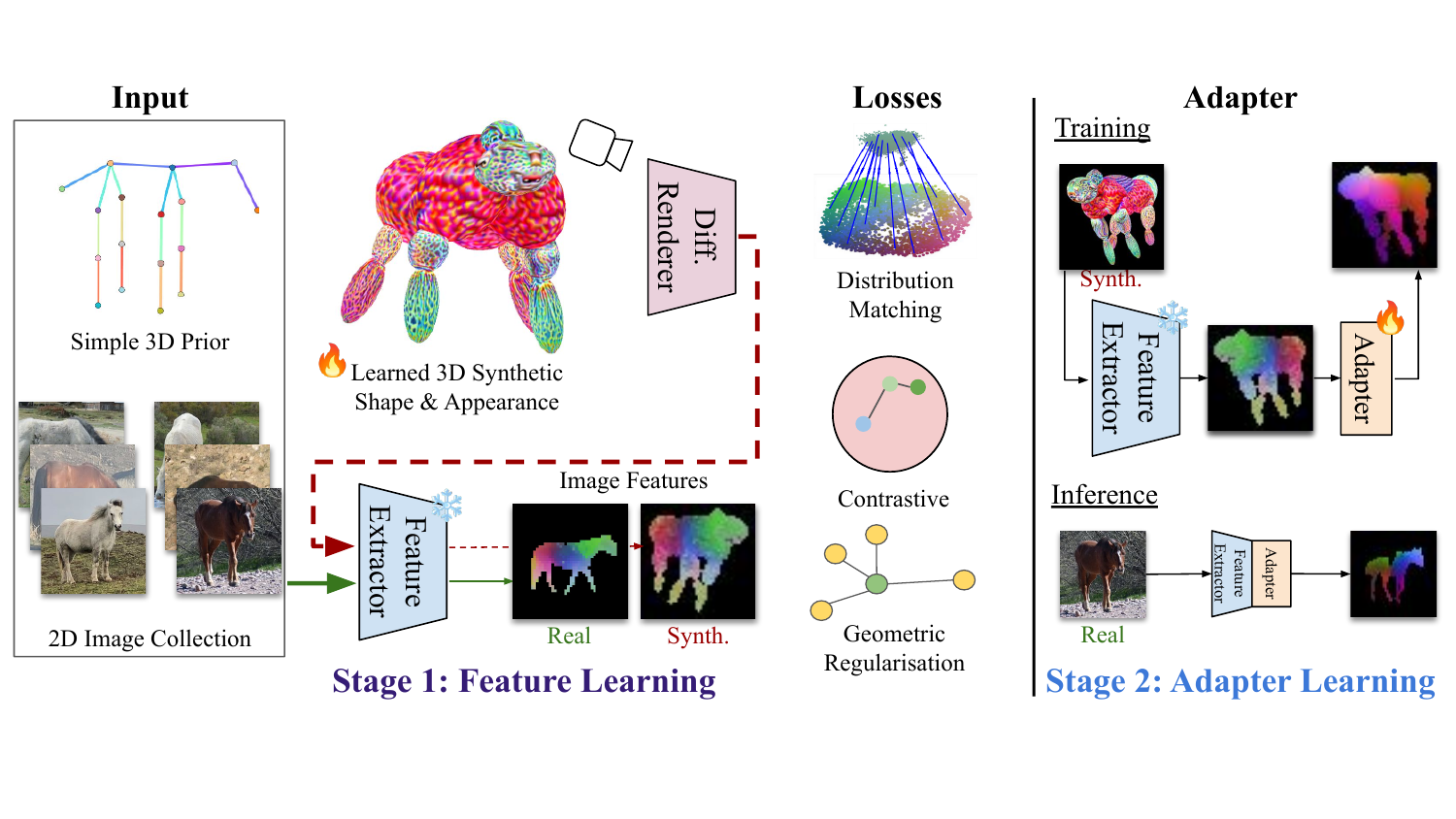}
\vspace{-30pt}
\caption{Overview of our approach which consists of feature learning via canonical shape reconstruction and then feature adapter learning. 
In the first stage, we start with a collection of images depicting the same object category (\eg a horse) and an initial simple 3D shape prior in the form of a graph. 
We then optimise the shape and  appearance of the canonical shape such that image features derived from it match the distribution of those obtained from real images, all without requiring any instance-level supervision. 
In the second stage, we refine our features using an adapter by exploiting the fact that in the first stage we have generated a mechanism for creating paired weakly supervised training data that maps image features to per-pixel canonical 3D coordinates. 
}
        \vspace{-15pt}
\label{fig:method}
\end{figure}

Given a dataset of real images $\{I_i\}_{i=1}^N$, depicting a category of interest, we want to find a mapping that maps each pixel on the object from one image onto the same semantic part in the other image. 
However, obtaining supervised correspondence annotations is time consuming and expensive, and thus we propose a fully self-supervised approach. 
Specifically, we take a pre-trained feature extractor $\phi$ as our starting point, and aim to refine the features it produces $\mathcal{F}=\phi(I) \in\mathbb{R}^{X\times Y\times D}$, with an adapter $\chi$ to disambiguate symmetric and repeated parts. 
The features we will use for matching are therefore $Q=\chi(\phi(I))$. 
The core idea of our approach is to learn a parameterised synthetic 3D object representation $R(\mathbf{x};\boldsymbol{\eta})$, for 3D positions $\mathbf{x}=(x,y,z)$ and articulation $\eta$, that is optimised to generate realistic features when 2D projections/renderings $\mathcal{P}_{v}(R)$ of it from camera view $v$  are passed to a pre-trained feature extractor $\phi$. 

The canonical representation $R(\mathbf{x};\eta)$ is a textured 3D mesh generated from a simple input graph that can be rendered to arbitrary views with arbitrary articulations $\eta$. 
By providing a 3D graph as input, depicting  a category's simplified skeleton topology in an approximately canonical pose, we construct the 3D representation by initialising 3D ellipsoids along each edge of the graph. 
The insight of Fig.~\ref{fig:motivation} is that realistic looking shapes are not required to obtain good features. 
We exploit this by using a simple and easy-to-manipulate representation.
The parameters being optimised then are the orthogonal scales $\boldsymbol{\sigma}^b=(\boldsymbol{\sigma}_x^b,\boldsymbol{\sigma}_y^b,\boldsymbol{\sigma}_z^b)$ for each $b^{th}$ ellipsoid,  the lengthwise scalar offset between the graph edge's 3D center and the ellipsoid's center  $\delta^b$, and each ellipsoid's texture.

We represent each ellipsoid as a 3D mesh with the texture parameterised as per-vertex RGB vectors, $\zeta_{i=1,...,N_w}^b$ where $\zeta_i^b=(r_i^b,g_i^b,b_i^b)$ and $N_w$ is the number of vertices per ellipsoid. 
As most real world object categories are symmetric about at least one plane, we enforce this symmetry by defining our graph to lie along the z-axis such that $R(x,y,z) = R(-x,y,z)$ meaning the number of parameters to be optimised is almost halved.  
We argue that with a suitable prior distribution for camera views, and for sufficiently many views, we should be able to optimise the canonical representation through self-supervised matching of synthetic and real feature distributions.
Once we have an optimised canonical representation where by construction we know the mapping from 2D image pixels to 3D positions on the canonical shape, we train an adapter $\chi$ to map from pre-trained features to improve semantic correspondence.

\vspace{-5pt}
\subsection{Feature Learning}
\vspace{-5pt}
We start with a pre-trained feature extractor $\phi$ that takes an image $I$ as input and produces a feature map $\mathcal{F} = \phi(I)$. 
These features are typically high-dimensional (\eg the dimensionality of DINOv2 patch tokens is 768)  and before we proceed we  reduce this dimensionality using Princicpal Components Analysis (PCA). 
We do this using only features from the category being optimised and a pre-trained model $\mathcal{M}$, \eg an off-the-shelf pre-trained object segmentation model such as SegmentAnything~\cite{kirillov2023segment}, to mask these features so only the relevant foreground ones are used. 
To match features from different images it is typical to find pairs that minimise cosine similarity.
If $\mathcal{F}=\phi(I)$ is the 2D grid of features representing an input image $I$, then the feature vector $\mathbf{f}$ located at pixel $(x,y)$ is denoted by $\mathcal{F}(x,y)$.
Therefore, following PCA we normalise all feature vectors as $\mathbf{f}=||\mathbf{f}||_2$ so we may optimise for the property we are interested in, \ie the  cosine similarity between real and synthetic features.

We seek to learn a 3D synthetic representation that replicates the spatial distribution of features produced by a pre-trained feature extractor on real images of the same category while avoiding the need for view or correspondence supervision. 
To do this we cannot rely on render-and-compare approaches like previous models such as~\cite{li_learning_2024, wu_magicpony_2023} do. 
Instead, we look to match the distributions of features in aggregate, independent of spatial structure. 
The underlying assumption being made is that for a suitable view prior distribution and for sufficiently many views, the distributions of the synthetic and real feature sets should be similar. 
More concretely, if we observe the category of interest from enough viewpoints we would expect the proportion of head features to leg features aggregated across all views to be approximately constant. 
That means we expect the aggregated histogram of all features across all views to be similar for both real images and synthetic renders.  

In learning a canonical representation, one approach would be to learn features assigned to the surface coordinates, similar to SHIC~\cite{leonardis_shic_2025}. But, as argued in \cite{zhang2024telling}, global pose information is contained within pre-trained features like DINOv2 and we argue an effective model should be able to extract more information from them by optimising a 3D source object from which to extract features, rather than optimising the features themselves on the 3D surface. 
By learning a representation that feeds into the feature extractor we obtain features that take advantage of the pre-learned spatial distributions of the extractor.

One of the motivations for our model is to consider uncommon non-trivial categories for which hand-crafted meshes, which are abundant for quadrupeds, are unavailable. 
We therefore design our model to be agnostic to the specific category as long as an approximate skeletal graph is provided as input. 
The graph is not expected to be carefully proportioned, \eg in our later experiments all edges have been made unit length. 
Additionally the input articulation, in which the graph is constructed, need only be approximate as it will be randomly perturbed during training.

\noindent\textbf{Distribution Matching}.
Given the aforementioned initial 3D shape for a category of interest we can render it from different virtual viewpoints and extract visual features from these images using a pre-trained vision backbone (\eg  DINOv2). 
However, despite the ability of methods like DINOv2 to generate semantically consistent visual features across different depictions of an object  (see Fig.~\ref{fig:motivation}),  these features will initially be  substantially different from those extracted from real images depicting the category. 
Thus the goal is to modify the shape and appearance  of the 3D template so that features extracted from renders of it are similar to those from real images.

There are several possible approaches for measuring the similarity between the two aggregate feature distributions, given that we do not have instance-level (\ie image-level) correspondences. 
Optimal transport-based approaches~\cite{peyre_computational_2018} attempt to find the least costly means of mapping samples from a source distribution to those in a target distribution according to some cost function $\mathbf{C}$, defining the distance between samples of the source distribution and samples of the target distribution
We choose optimal transport in contrast to alternative approaches such as Maximum Mean Discrepancy (MMD) \cite{scholkopf_kernel_2007} as, in its regularised form, it is as computationally efficient as MMD~\cite{peyre_computational_2018} but we also obtain a mapping for each of our synthetic feature samples to real feature samples which we will use at a later stage.

In optimal transport we seek to optimise a transport map $\mathbf{P}$, which assigns probability mass from one distribution to another, by minimising its inner product with the cost function or matrix $\mathbf{C}$. 
The cost matrix elements $\mathbf{C}_{ij}$ give the distance from source sample $i$ to target sample $j$. 
In the discrete case, once optimised, the row $\mathbf{P}_{i,:}$ gives the transport of the $i^{th}$ sample from the source distribution to each of the target samples. 
In an unregularised empirical sampling setting there would be one non-zero entry per row. 
A discrete measure for each distribution constrains the optimisation requiring each row and column of $\mathbf{P}$ sum to each sample's total weight via $\mathbf{P}\mathbbm{1}_m = \mathbf{a}$ and $\mathbf{P}^T\mathbbm{1}_n=\mathbf{b}$, where n and m are the number of source and target samples, and $\mathbf{a}$ and $\mathbf{b}$ are vectors of length equal to the number of samples $n$ and $m$ respectively with each entry having value $\frac{1}{n}$ and $\frac{1}{m}$  for an empirical measure.
The entropy, $\mathbf{H}$, regularised discrete optimal transport problem  with regularisation parameter $\epsilon$, and set of permissible transport maps $\mathbf{U}(\mathbf{a},\mathbf{b})$, can be written in the Kantorovich formulation~\cite{peyre_computational_2018}:

\begin{equation}
\mathcal{L}_{\epsilon}(\mathbf{a}, \mathbf{b}) = \min\limits_{\mathbf{P}\in\mathbf{U}(\mathbf{a}, \mathbf{b})}\sum\limits_{i,j} \mathbf{C}_{ij}\mathbf{P}_{ij} - \epsilon \mathbf{H}(\mathbf{P}), 
\label{eq-kantorovich}
\end{equation}

\begin{equation}
\mathbf{H}(\mathbf{P}) = \sum\limits_{i,j} \mathbf{P}_{ij}(log\mathbf{P}_{ij}-1), 
\end{equation}

\begin{equation}
\mathcal{W}_{\epsilon}(\mathbf{a}, \mathbf{b}) = \sum\limits_{i,j} \mathbf{C}_{ij}\mathbf{P}_{ij}^*, 
\end{equation}
where, $\mathbf{P}^*$ is our optimal transport and $\mathcal{W}_{\epsilon}$ is the Wasserstein distance. 
Unfortunately the gradients calculated for the Wasserstein distance between discrete sample sets are biased \cite{bellemare_cramer_2017} but this can be remedied by using instead the Sinkhorn Divergence; the energy distance with the Wasserstein distance as metric, together with entropic regularisation  \cite{salimans_improving_2018, genevay_learning_2018}: 

\begin{equation}
\mathcal{L}_{distr} (\mathbf{a},\mathbf{b}) = 2 \mathcal{W}_{\epsilon} (\mathbf{a},\mathbf{b}) - \mathcal{W}_{\epsilon} (\mathbf{a},\mathbf{a}) - \mathcal{W}_{\epsilon} (\mathbf{b},\mathbf{b})
\label{eq-sinkh-loss}
\end{equation}
This linear program is optimised using the Sinkhorn algorithm~\cite{peyre_computational_2018}.

\noindent\textbf{Contrastive Learning}. 
Distribution matching is not the only means of optimising the synthetic features. Self-supervised optimisation of the shape and texture can also be achieved through geometric consistency.
To obtain a representation that is consistent across a large variety of arbitrary views and one that captures the full range of features we are likely to see for real images we want to enforce smooth variation in features corresponding to nearby points on the 3D surface and to pull similar features together on the surface to form coherent parts. 
To do this we exploit our underlying 3D geometry and use contrastive losses on the features back-projected onto the vertices of the canonical representation from the 2D features.

For a given camera view $v$ from a set of $\mathcal{N}_V$ views we randomly sample $\mathcal{N}_S$ vertices from our mesh together with their corresponding features $\mathbf{f}_s^v$ which are those back-projected from the features onto the vertices. 
These are our anchor features in the contrastive loss. 
We merge our ellipsoids into one 3D shape by culling those vertices appearing inside other ellipsoids and greedily adding edges from each vertex on the newly formed boundaries to the nearest vertex in the neighbouring ellipsoid. 
Here the neighbouring ellipsoid is the parent or child in the original 3D graph. 
With this fully connected graph and starting from each anchor vertex we find all vertices that are within $r$ edges of the anchor. 
The corresponding features are the anchor's positive set from which we can randomly sample $\mathcal{N}_a$ features $\{\mathbf{f}_a^v\}$. 
The complement of this inner circle of vertices, \ie the rest of the graph, is the anchor's negative set and we randomly sample an equal number of $N_a$ features $\{\mathbf{f}_b^v\}$ to give, with constant $\alpha$:  

\begin{equation}
\mathcal{L}_{loc}(\mathbf{f}_s,\mathbf{f}_a,\mathbf{f}_b) = \frac{1}{N_a}\sum\limits_{n}^{N_a} max(0, -\mathbf{f}_s^T\mathbf{f}_{a,n} + \mathbf{f}_s^T\mathbf{f}_{b,n} + \alpha),
\end{equation}
\begin{equation}
\mathcal{L}_{contr} = \frac{1}{N_VN_s}\sum\limits_v^{N_V}\sum\limits_i^{N_s}\mathcal{L}_{loc}(\mathbf{f}_{s_i}^{v}, \mathbf{f}_{a_i}^{v}, \mathbf{f}_{b_i}^{v}). 
\label{eq-contrastive-loss}
\end{equation}

\vspace{-5pt}
\subsection{Geometric Consistency}
\vspace{-5pt}
One of the challenges with learning from unlabelled images, whether we are using distribution matching or  the contrastive loss, is it is difficult to determine which images are the front and which are the back and where the lines of symmetry should be. 
As a consequence it is easy for the model to learn, for example, two heads for an animal as images centered on the face are more common in biased datasets. 
We try to mitigate this by enforcing view-consistency. For both contrastive and distribution matching algorithms we minimise the variance at each vertex of visible features assigned to that vertex. 
We do this by minimising one minus the cosine similarity of visible vertex features with the vertex's mean feature $\mathbf{\bar{f}}$: 

\begin{equation}
\mathcal{L}_{var} = \frac{1}{N_w}\sum_w^{N_w} 1 - \mathbf{f}_{w,vis}^T\mathbf{\bar{f}}_{w}. 
\label{eq-var-loss}
\end{equation}

To further exploit our geometry we use the symmetry defined for our category graph and minimise the cosine similarity of features on symmetric vertices, \ie enforcing that left and right should be equivalent: 
\begin{equation}
\mathcal{L}_{sym} = \frac{1}{N_{sym}}\sum\limits_{i}^{N_{sym}} \mathbf{f}_i^T\mathbf{f}_{sym(i)},
\label{eq-symmetric-loss}
\end{equation}
where for vertex index $i$, $sym(i)$ indicates the index of its symmetric vertex. The total number of unique vertices, excluding symmetries is $N_{sym}$. Again this is done for both learning algorithms.

Our final geometric loss term comprises $\mathcal{L}_{var}$, which optimises the consistency across views, and $\mathcal{L}_{sym}$ which encourages symmetry:  
\begin{equation}
\mathcal{L}_{geoc} = \mathcal{L}_{var} + \mathcal{L}_{sym}.
\label{eq-geometric-loss}
\end{equation}

\noindent\textbf{Geometric Regularisation}. 
In addition to the above losses, we add geometric regularisation terms. 
Specifically, we add an L1 loss on the scales $\sigma$ and offsets $\delta$ (relative to their original positions at 0.5), a collision penalty, $\mathcal{L}_{coll}$, where non-neighbouring ellipsoids are penalised for intersecting, and a texture smoothing term $\mathcal{L}_{\zeta}$. 
The collision penalty takes the line between non-neighbouring ellipsoid centers, finds the vertices on each ellipsoid whose vector from their respective ellipsoid center is most closely aligned with this line, and uses the distance between their projected positions on the line as a measure of overlap. 
The texture regularisation is an L1 difference between neighbouring vertex texture parameters $\zeta$. The complete regularisation therefore takes the form:

\begin{equation}
\mathcal{L}_{reg} = \lambda_{\sigma}\sum|\sigma| + \lambda_{\delta} \sum|\delta-0.5| + \lambda_c\mathcal{L}_{coll} + \lambda_{\zeta}\mathcal{L}_{\zeta}, 
\end{equation}
which combines the following terms:  
\begin{equation}
\mathcal{L}_{coll} = \sum_i \sum_j \mathbbm{1}_{\neg neighs(i,j)} \text{ReLU}(\max_a [(c_i - c_j)^Tw_a^j] - \min_b[(c_i-c_j)^Tw_b^i]),  
\end{equation}

\begin{equation}
\mathcal{L}_{\zeta} = \frac{1}{N_{edges}}\sum_{i,j} |\zeta_i - \zeta_j|\mathbbm{1}_{edge(i,j)}. 
\end{equation}

The final combined loss function that we optimise is:

\begin{equation}
\mathcal{L} = \mathcal{L}_{X} + \mathcal{L}_{geoc} + \mathcal{L}_{reg}.
\label{eq-total-loss}
\end{equation}
where $\mathcal{L}_X$ is $\mathcal{L}_{distr}$ or $\mathcal{L}_{contr}$.
While the model is able to learn features that are semantically similar to real ones, the constraints of the representation prevent it from being able to replicate them perfectly. 
To bridge that final gap we map these synthetic features in minibatches to features found in real features using optimal transport, as in the Distribution Matching section.

\vspace{-5pt}
\subsection{Feature Adapter}
\vspace{-5pt}

Having learned a 3D representation that can reconstruct the features present in real images and structure them spatially in 3D, we can now use this representation to learn features for arbitrary camera views for the category of interest. 
Effectively, we now have the ability to generate paired data consisting of visual features with their respective per-pixel 3D locations on our learned 3D representation by rendering the 3D shape and passing the resulting images through the pre-trained feature extractor $\phi$. 
We can exploit the learned 3D structure of our distribution of features to learn new features possessing this geometric information. 
Given this paired data we can train a feature adapter that takes features $\mathcal{F}$ as input and maps them to new features corresponding to predictions of the 3D canonical positions, $Q$, on the representation for each image patch, \ie $Q = \chi(\mathcal{F})$, along with a predicted mask $\mathbf{M}_Q$. 
The learned 3D representation is not guaranteed to be geometrically similar to the category thus we rely on the adapter to learn the relative spatial pattern of features. 
To train this adapter we adopt an approach similar to DualPM \cite{kaye_dualpm_2025} and DUSt3R~\cite{wang_dust3r_2024} by regularising an L2 prediction loss with a per-pixel confidence prediction $c_Q$~\cite{kendall2017uncertainties}: 

\begin{equation}
\mathcal{L}_{adp} = \frac{1}{|\mathbf{M}_Q|}\sum\limits_{u}c_{Q}(u)||\hat{Q}_u - Q_u ||^{2} - \lambda_c log(c_{Q}(u)), 
\end{equation}
where $\mathbf{M}_Q$ is the predicted mask, $\hat{Q}_u$ and $Q_u$ are the ground truth and predicted position maps for masked feature tokens $u$, $c_Q$ is the predicted confidence map, and $\lambda_c$ is a hyperparameter. 
The predicted mask is trained using binary cross entropy with the known synthetic mask.

Once $\chi$ has been trained, we can then predict the 3D position on our canonical 3D representation for real images by first extracting their features using the pre-trained feature extractor $\phi$, and then applying the adapter to get $Q = \chi(\phi(I))$. 
These 3D positions act as our features as they should correspond each original backbone feature $\mathbf{f}$ with a unique position on the representation. 
For finding correspondences between pixel locations on two different images $I_1$ and $I_2$ with corresponding features $\mathcal{F}^1$ and $\mathcal{F}^2$, we use a squared Euclidean metric, $d(\mathbf{f}_i^1,\mathbf{f}_j^2) = || \chi(\mathcal{F}^1)_i - \chi(\mathcal{F}^2)_j ||^{2}$ where $i$ and $j$ signify the $i^{th}$ and $j^{th}$ feature tokens in each map corresponding to the two pixel locations

Finally, when performance matching, we can exploit the complementary advantages of the original pre-trained features and our 3D ones by combining them. 
First we can remove features using the predicted mask $\mathbf{M}_Q$.
If we then find the top k matches using the original pre-trained features, we can refine the selection using the closest match among these $k$ matches from our model's predictions: 

\begin{equation}
d_{tk}(\mathbf{f}_i^1,\mathbf{f}_j^2) =
\begin{cases}
|| \chi(\mathcal{F}^1)_i - \chi(\mathcal{F}^2)_j ||^{2} & \text{if } \text{TopK}((\mathbf{f}_i^1)^T\mathbf{f}_j^2,k)\\
\infty & \text{otherwise}. 
\end{cases}
\end{equation}

\vspace{-5pt}
\section{Results}
\vspace{-5pt}
Here we report both quantitative and qualitative results across several different object categories and different ablated variants of our model. 
Implementation and training details can be found in the appendix.

\vspace{-5pt}
\subsection{Quantitative Results}
\vspace{-5pt}

We evaluate our model on the task of semantic correspondence estimation using the PCK@0.1 metric on PF-Pascal~\cite{ham_proposal_2018} for different quadruped categories. 
Results presented in Table~\ref{table:pascal_results}. 
We evaluate on the two template learning approaches: (i) contrastive with contrastive loss on the synthetic features without distribution matching (Eqn.~\ref{eq-sinkh-loss}) and (ii) distributive, global distribution matching without the contrastive loss. The results show the optimisation is successful with either algorithm. 
PF-Pascal is a commonly used benchmark and allows us to compare to other similar models. We first compare to SphericalMaps~\cite{mariotti_improving_2024} using the protocol and test split set out in~\cite{zhang2024telling}. This model interpolates its final features between its spherical features and the original DINOv2 features using a interpolative factor $\alpha$. We compare to $\alpha=0.8$, the default value, and $\alpha=0.0$ which is just the sphere features prediction, most comparable to our direct results. Our model performs comparably across all four categories despite no view labels being used.
We then compare our model to the set of self-supervised reconstruction models, see Table~\ref{table:pascal_voc_results}. Note that the image pairs used to evaluate these are no longer available and so the comparison is not exact. 

To try to recreate the protocol used in~\cite{kaye_dualpm_2025, kulkarni2020articulation}, we sampled 10k pairs of images from the PF-Pascal full dataset and cropped around the bounding box of each instance of a category in the image. The cropped images were then resized to 256$\times$256 and keypoint transfer calculated. The original bounding box was scaled by the crop scaling and 0.1 times the scaled bounding box used as the PCK threshold.
Our model out performs each of the self-supervised methods MagicPony and Farm3D, self-supervised models explicitly designed to reconstruct quadrupeds. though does not quite match the highly-supervised DualPM model~\cite{kaye_dualpm_2025}.
Observing the test images, we find the topk model matches features DINOv2 was unable to. 
In the appendix we show additional qualitative examples, showing that, for the cow example, the model is able to match both eyes while DINOv2 matched only one, and was able to better match the legs.
This improvement in left-right disambiguation is quantified in Table~\ref{table:symmetry_pck} where we report superior matching performance on keypoints in AwAPose that have explicit left and right confusion (\eg left versus right front leg). 

\begin{table}[t]
  \centering
  \small
  \begin{minipage}{0.56\linewidth}
    \centering
    \resizebox{1.0\textwidth}{!}{
    \begin{tabular}{lccc}
      \toprule
      & \Cow & \faIcon{horse} & \Sheep\\ 
      \midrule
      DINOv2~\cite{oquab_dinov2_2023} & 87.5 & 84.3 & 80.0  \\
      DINOv2~\cite{oquab_dinov2_2023} PCA16 & 83.3 & 85.5 & 80.0  \\ \hline
      SphericalMaps~\cite{mariotti_improving_2024} (sphere only) & 64.6 & 61.5 & 80.0 \\
      SphericalMaps~\cite{mariotti_improving_2024} & 91.7 & 85.5 & 80.0 \\
 \hline
      Ours Contrastive & 81.6 & 73.5 & 60.0 \\
      Ours Distributive & 83.3 & 72.3 & 60.0 \\ %
      Ours Contrastive (with topk)  & 97.9 & 80.7 & 80.0 \\
      Ours Distributive (with topk) & 93.8 & 84.3 & 80.0 \\
      \bottomrule
    \end{tabular}
    }
  \end{minipage}
  \hfill
  \begin{minipage}{0.42\linewidth}
    \caption{
      Evaluation on PF-Pascal using PCK@0.1 for cosine similarity nearest neighbour from source to       target. 
      We follow other papers and use the image resolution when
      calculating the threshold rather than the bounding box. For topk, we used k=5 matches.
    }
    \label{table:pascal_results}
  \end{minipage}
  \vspace{-10pt}
\end{table}

\begin{table}[t]
  \centering
  \small
  \begin{minipage}{0.50\linewidth}
    \centering
    \resizebox{1.0\textwidth}{!}{
    \begin{tabular}{lccc}
      \toprule
      & \Cow & \faIcon{horse} & \Sheep\\ 
      \midrule
      DINOv2~\cite{oquab_dinov2_2023} & 53.4 & 59.8 & 51.1 \\
      DINOv2~\cite{oquab_dinov2_2023} PCA16 & 51.8 & 59.8 &  50.7 \\ \hline
      A-CSM~\cite{kulkarni2020articulation} & 26.3 & 32.9 & 28.6 \\
      MagicPony~\cite{wu_magicpony_2023} & 42.5 & 42.9 & 41.2 \\
      Farm3D~\cite{jakab_farm3d_2024} & 40.2 & 49.1 & 36.1 \\
      DualPM\cite{kaye_dualpm_2025} (supervised) & 66.9 & 73.2 & 66.8 \\
 \hline
      Ours Contrastive & 46.7 & 51.0 & 34.7 \\
      Ours Distributive & 43.9 & 49.9 & 28.6 \\
      Ours Contrastive (with topk) & 57.1 & 60.5 & 54.3\\
      Ours Distributive (with topk) & 56.2 & 61.9 & 53.3 \\
      \bottomrule
    \end{tabular}
    }
  \end{minipage}
  \hfill
  \begin{minipage}{0.48\linewidth}
    \caption{
      Keypoint matching on PF-Pascal using PCK@0.1. %
      A-CSM, MagicPony, Farm3D, and DualPM project their 3D reconstruction vertices onto each image and assign the source feature to the nearest projected vertex which is then matched to the feature it is projected onto in the target image. 
      Like~\cite{kaye_dualpm_2025,wu_magicpony_2023}, we use the image resolution when calculating the threshold rather than the bounding box. For topk, we  used k=5 matches.
    }
    \label{table:pascal_voc_results}
  \end{minipage}
  \vspace{-10pt}   
\end{table}

\begin{figure}[h]
    \centering
    \begin{minipage}{0.3\textwidth}
        \centering
        \includegraphics[width=0.4\linewidth]{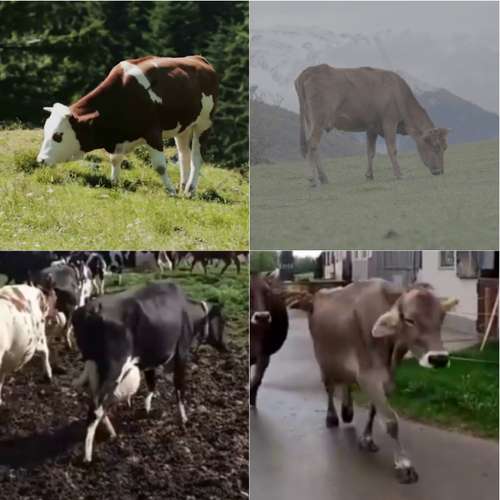}
        \includegraphics[width=0.4\linewidth]{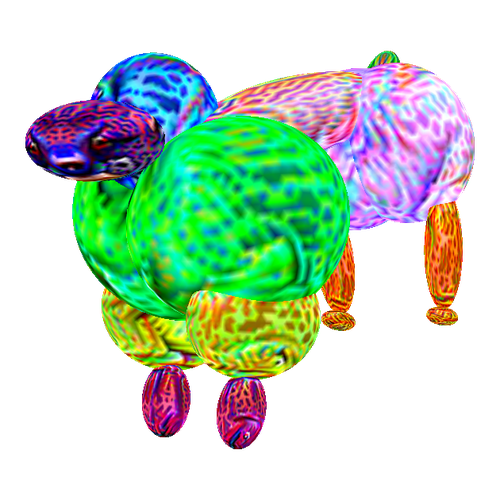}
        \\
        \includegraphics[width=0.4\linewidth]{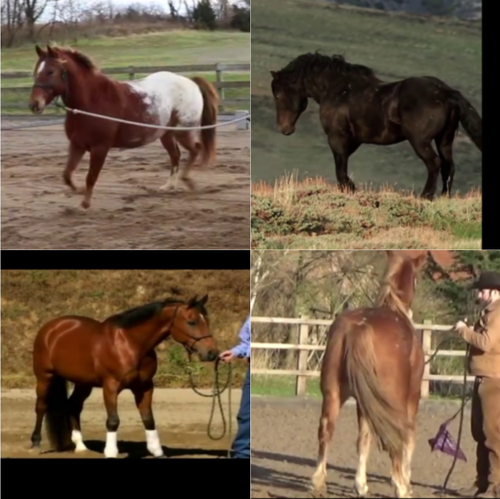}
        \includegraphics[width=0.4\linewidth]{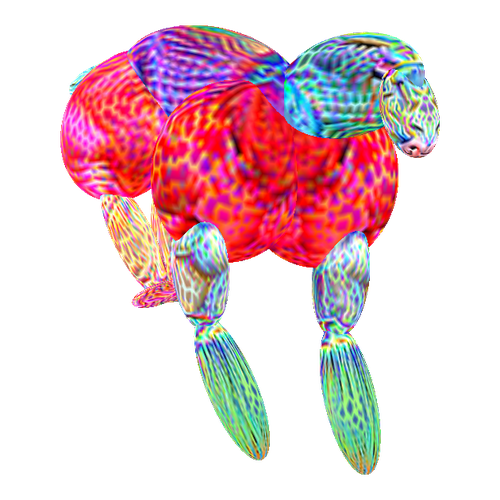}
        \\
        \includegraphics[width=0.4\linewidth]{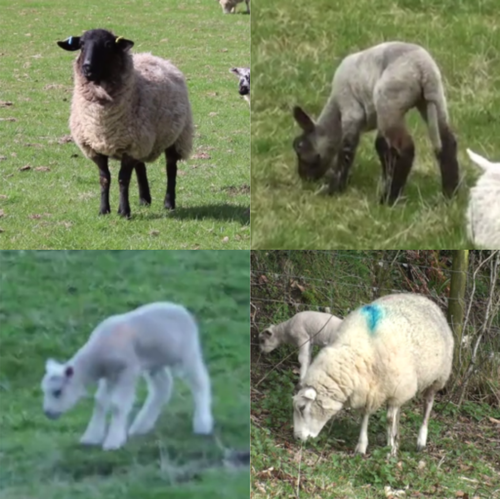}
        \includegraphics[width=0.4\linewidth]{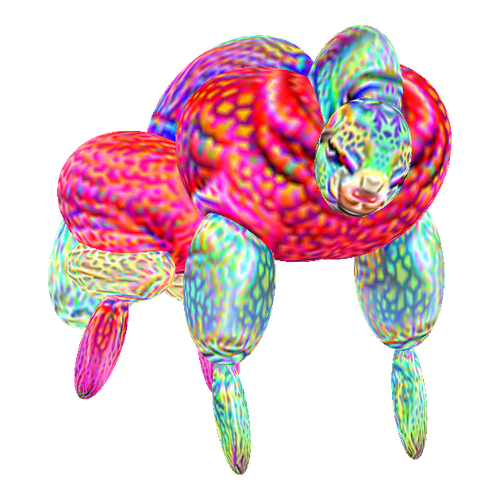}
        \\
        \includegraphics[width=0.4\linewidth]{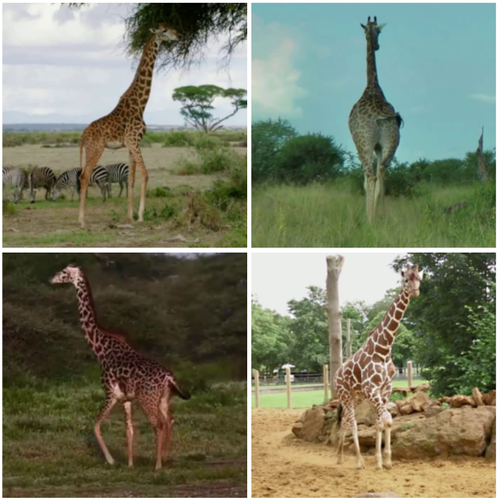}
        \includegraphics[width=0.4\linewidth]{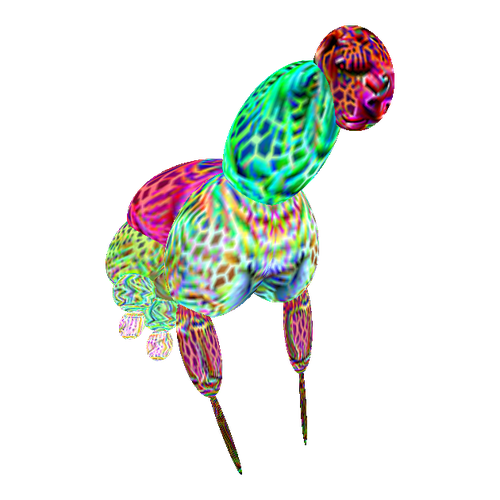}
        \\
        \includegraphics[width=0.4\linewidth]{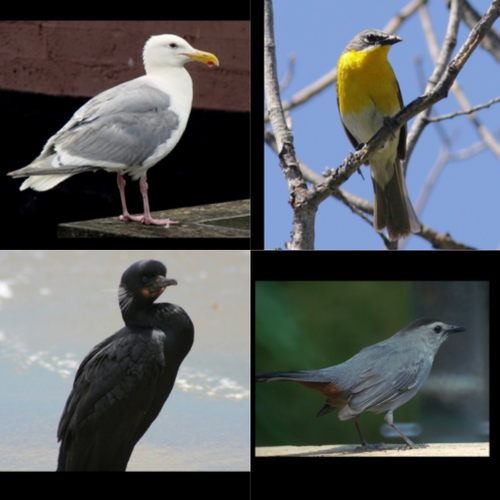}
        \includegraphics[width=0.4\linewidth]{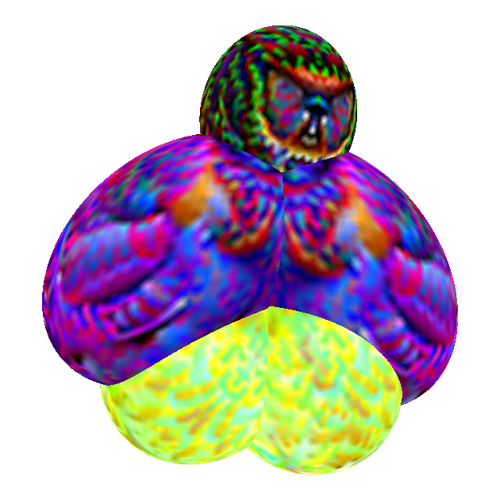}
        \\
        \includegraphics[width=0.4\linewidth]{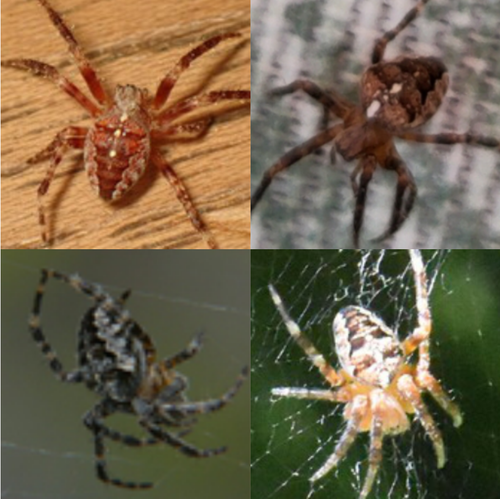}
        \includegraphics[width=0.4\linewidth]{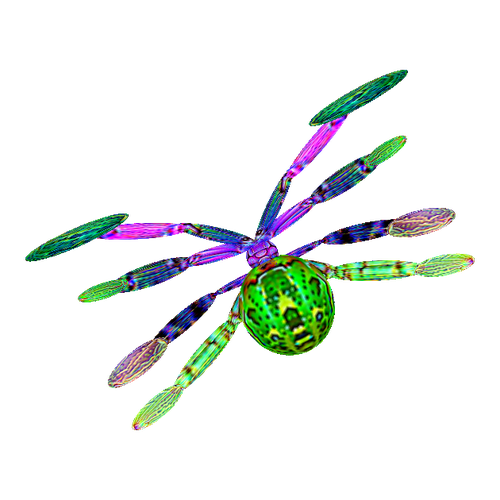}
        \\
        \includegraphics[width=0.4\linewidth]{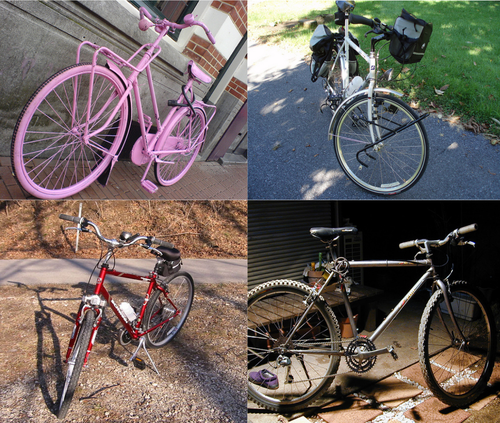}
        \includegraphics[width=0.4\linewidth]{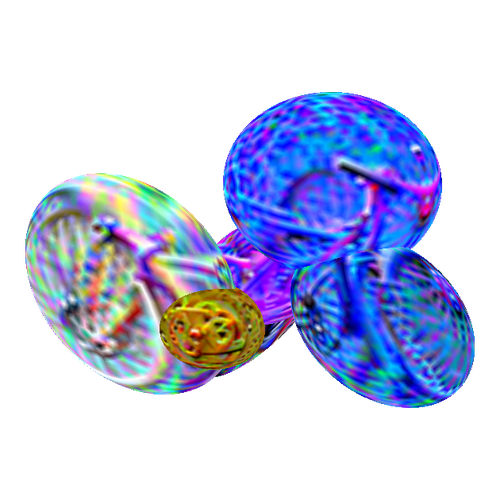}
    \end{minipage}
    \begin{minipage}{0.3\textwidth}
        \centering
        \includegraphics[width=0.4\linewidth]{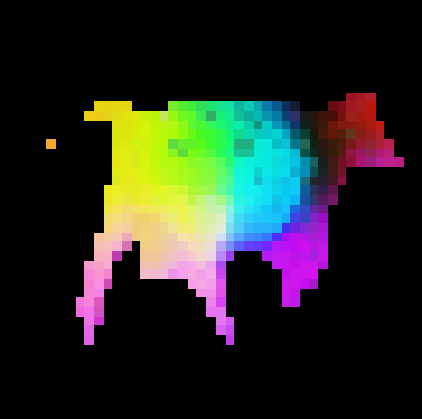}
        \includegraphics[width=0.4\linewidth]{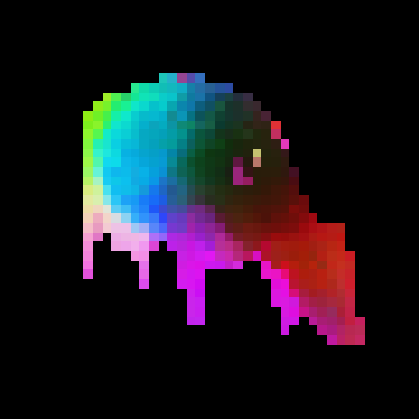}
        \\
        \includegraphics[width=0.4\linewidth]{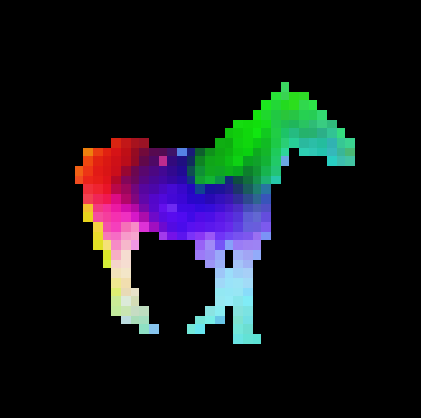}
        \includegraphics[width=0.4\linewidth]{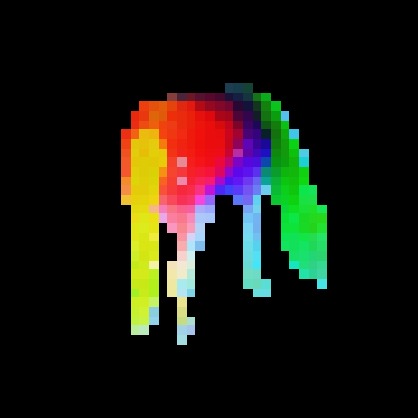}
        \\
        \includegraphics[width=0.4\linewidth]{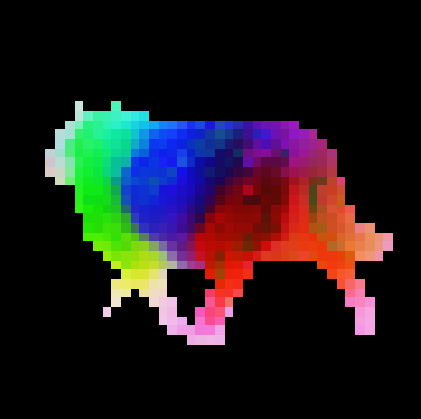}
        \includegraphics[width=0.4\linewidth]{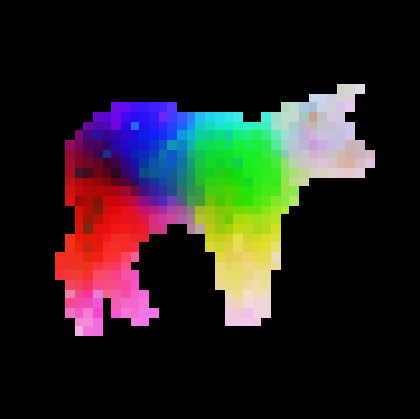}
        \\
        \includegraphics[width=0.4\linewidth]{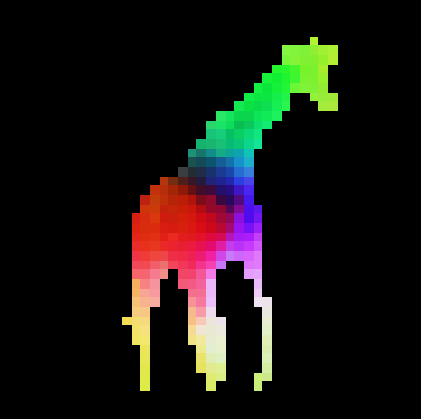}
        \includegraphics[width=0.4\linewidth]{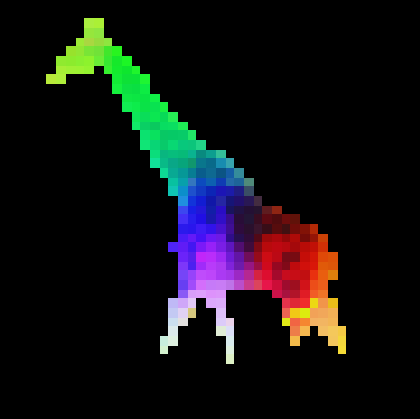}
        \\
        \includegraphics[width=0.4\linewidth]{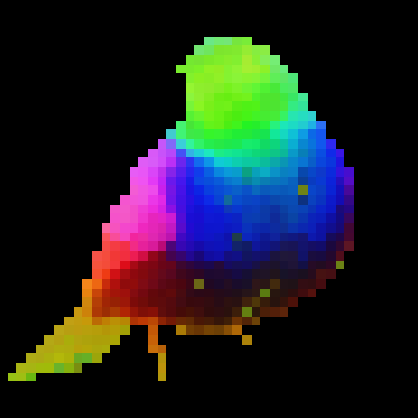}
        \includegraphics[width=0.4\linewidth]{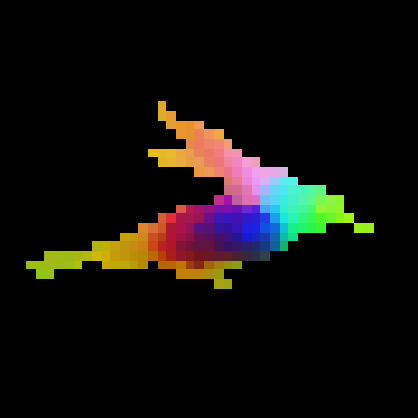}
        \\
        \includegraphics[width=0.4\linewidth]{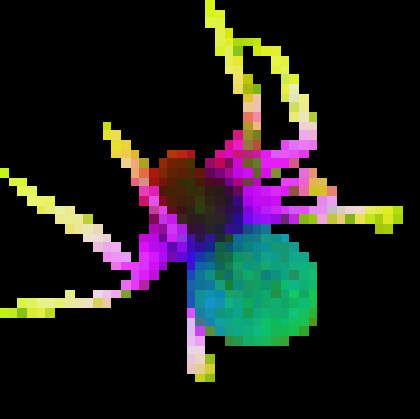}
        \includegraphics[width=0.4\linewidth]{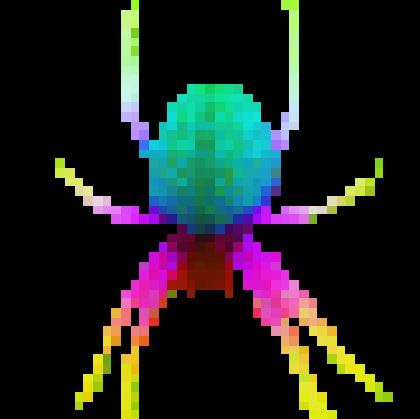}
        \\
        \includegraphics[width=0.4\linewidth]{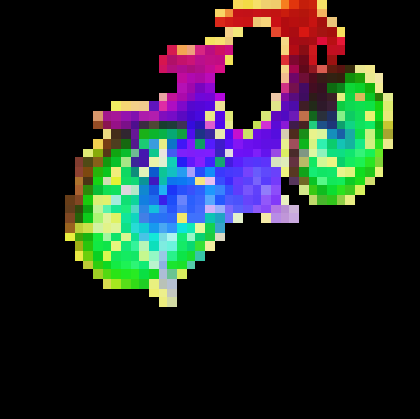}
        \includegraphics[width=0.4\linewidth]{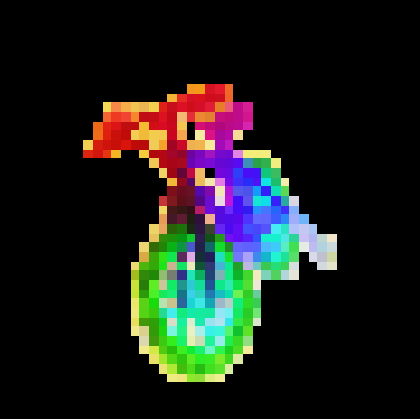}
    \end{minipage}
    \begin{minipage}{0.3\textwidth}
        \centering
        \includegraphics[width=0.4\linewidth]{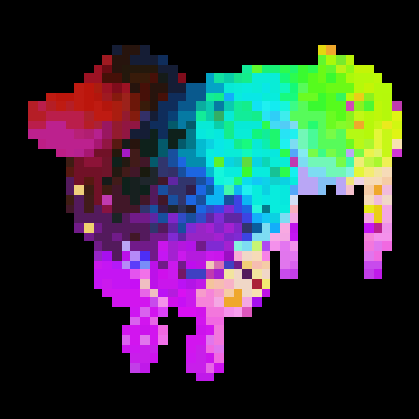}
        \includegraphics[width=0.4\linewidth]{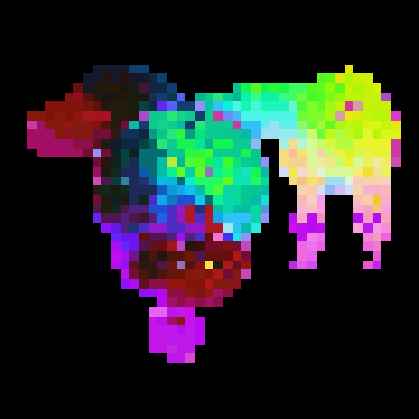}
        \\
        \includegraphics[width=0.4\linewidth]{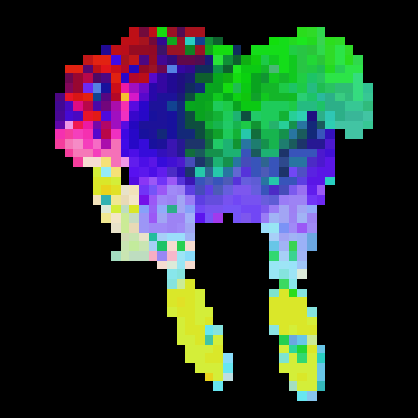}
        \includegraphics[width=0.4\linewidth]{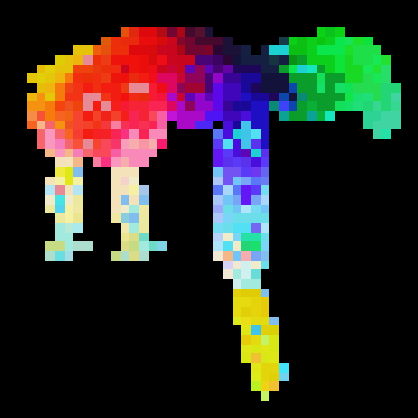}
        \\
        \includegraphics[width=0.4\linewidth]{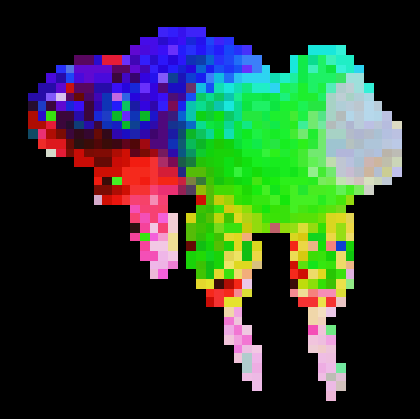}
        \includegraphics[width=0.4\linewidth]{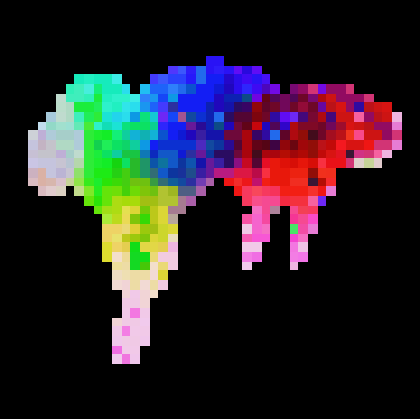}
        \\
        \includegraphics[width=0.4\linewidth]{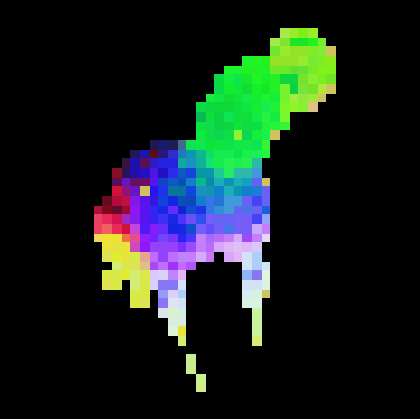}
        \includegraphics[width=0.4\linewidth]{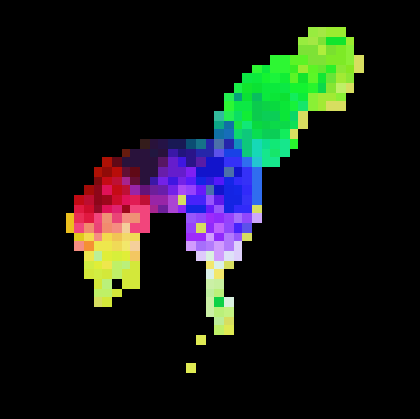}
        \\
        \includegraphics[width=0.4\linewidth]{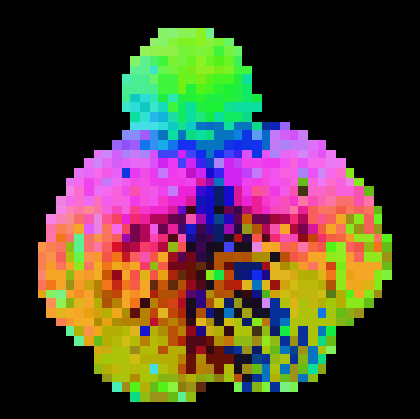}
        \includegraphics[width=0.4\linewidth]{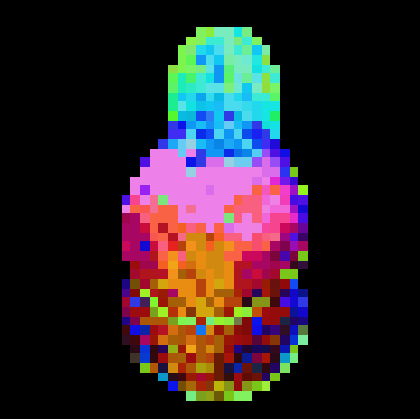}
        \\
        \includegraphics[width=0.4\linewidth]{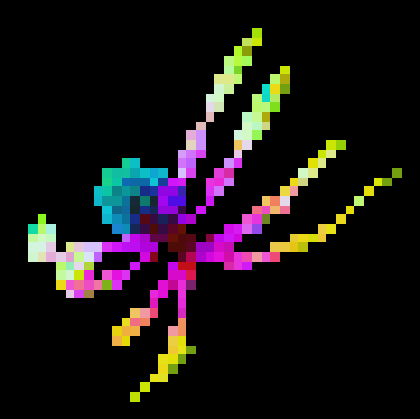}
        \includegraphics[width=0.4\linewidth]{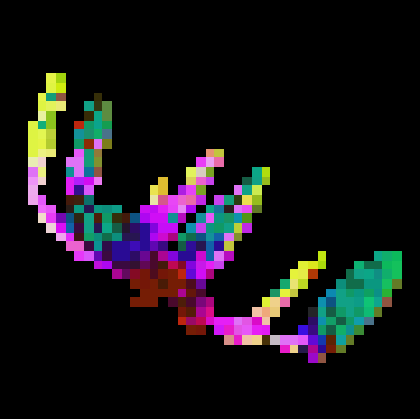}
        \\
        \includegraphics[width=0.4\linewidth]{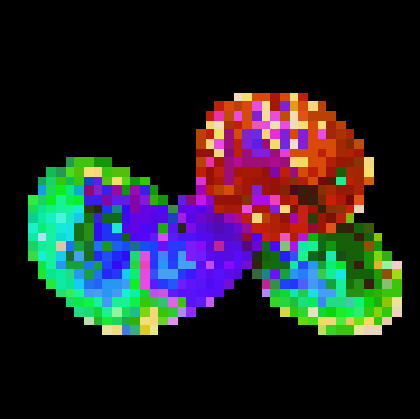}
        \includegraphics[width=0.4\linewidth]{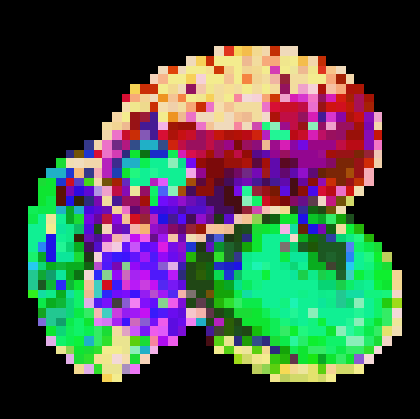}
    \end{minipage}
        \vspace{-3pt}
        \caption{Here we display the learned canonical 3D shape for different object categories using the Distributional approach. The first column shows representative training images for each category while the second column shows our learned canonical shape. The next two columns show DINOv2 features extracted from real images while the final two columns show DINOv2 features extracted for our synthetic images. The features are visualised using PCA. 
        Our resulting features are trained to look like those from real images, but the appearance of the 3D shape is not constrained to look realistic. 
        }
        \vspace{-10pt}
    \label{fig:cow_horse_sheep_synthetic_canonical_models}
\end{figure}

\vspace{-5pt}
\subsection{Qualitative Results}
\vspace{-5pt}

There are two stages of our model to examine qualitatively. 
There is the 3D canonical model and its spatial distribution of features and  there is the 3D position predictions from the adapter.
We are aiming to learn a representation that can capture the total range of features present in real images and replicate their 3D spatial distribution such that they are sensibly placed and as consistent as possible across all views.

In Fig.~\ref{fig:cow_horse_sheep_synthetic_canonical_models} we display reconstruction results for different categories with varying topologies. 
The first three all share the exact same intialisation (\ie the same input graph is provided) but develop into unique representations. 
In particular, each representation learns to develop a face on the texture that captures characteristics of its category so as to reconstruct features found in the dataset. 
The versatility of our approach enables feature reconstruction for categories that are commonly ignored by other models, \eg spiders and bikes.

\begin{figure}[t]
    \centering

    \begin{minipage}{0.32\textwidth}
        \centering
        \includegraphics[width=0.4\linewidth]{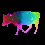}
        \includegraphics[width=0.4\linewidth]{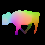}
        \\
        \includegraphics[width=0.4\linewidth]{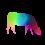}
        \includegraphics[width=0.4\linewidth]{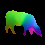}
        \\
        \includegraphics[width=0.4\linewidth]{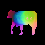}
        \includegraphics[width=0.4\linewidth]{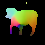}
        \\
        \includegraphics[width=0.4\linewidth]{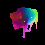}
        \includegraphics[width=0.4\linewidth]{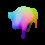}
        \\
        \includegraphics[width=0.4\linewidth]{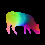}
        \includegraphics[width=0.4\linewidth]{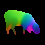}
    \end{minipage}
    \begin{minipage}{0.32\textwidth}
        \centering
        \includegraphics[width=0.4\linewidth]{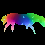}
        \includegraphics[width=0.4\linewidth]{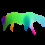}
        \\
        \includegraphics[width=0.4\linewidth]{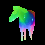}
        \includegraphics[width=0.4\linewidth]{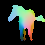}
        \\
        \includegraphics[width=0.4\linewidth]{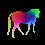}
        \includegraphics[width=0.4\linewidth]{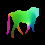}
        \\
        \includegraphics[width=0.4\linewidth]{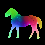}
        \includegraphics[width=0.4\linewidth]{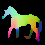}
        \\
        \includegraphics[width=0.4\linewidth]{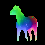}
        \includegraphics[width=0.4\linewidth]{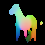}
    \end{minipage}
    \begin{minipage}{0.32\textwidth}
        \centering
        \includegraphics[width=0.4\linewidth]{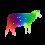}
        \includegraphics[width=0.4\linewidth]{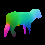}
        \\
        \includegraphics[width=0.4\linewidth]{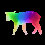}
        \includegraphics[width=0.4\linewidth]{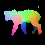}
        \\
        \includegraphics[width=0.4\linewidth]{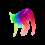}
        \includegraphics[width=0.4\linewidth]{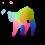}
        \\
        \includegraphics[width=0.4\linewidth]{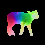}
        \includegraphics[width=0.4\linewidth]{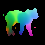}
        \\
        \includegraphics[width=0.4\linewidth]{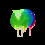}
        \includegraphics[width=0.4\linewidth]{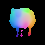}
    \end{minipage}
        \vspace{-3pt}
        \caption{Qualitative results of applying our adapter to DINOv2 features from real images using shapes learned using the Distributional approach. 
        Columns from left to right, depict results for cows, horses, and sheep, where each row is a different image. 
        For each column, the original DINOv2 features are on the left and the predicted 3D positions on the template representation are on the right. 
        Our results represent the predicted 3D coordinates mapping to a location on the learned template. 
        Unlike the original DINOv2 features, which in many instances are unable to encode left-right disambiguation, our model much more clearly separates such parts.}
        \vspace{-10pt}
    \label{fig:qualitative_examples}
\end{figure}

For the adapted features, we are primarily interested in whether we can discern semantic information that pre-trained features like DINOv2 struggle with, in particular telling left from right and, for animals, identifying each of the legs.
Results for the refinement adapter are shown in Fig.~\ref{fig:qualitative_examples} for three categories: cows, horses, and sheep. 
The adapted features for all three categories display a strong refinement of the original features, partially disambiguating left from right, with the various legs often differentiated.  
One of the issues we found for many categories is that our model can sometimes confuse a tail with a head or merges the tail with the back legs, a problem that can occur for real images too.

\vspace{-5pt}
\subsection{Ablations}
\vspace{-5pt}

\begin{table}[t]
  \centering
  \small
  \begin{minipage}{0.50\linewidth}
    \centering
    \resizebox{0.95\textwidth}{!}{
    \begin{tabular}{lrr}
      \toprule
      & PFP. & AwA \\ 
      \midrule
      Initial 3D Prior & 7.2 & 1.8 \\
      $\mathcal{L}_{contr} + \mathcal{L}_{geoc}$ & 37.3 & 16.0 \\
      $\mathcal{L}_{distr} + \mathcal{L}_{geoc}$ & 31.3 & 13.8 \\
      $\mathcal{L}_{contr} + \mathcal{L}_{geoc} + \mathcal{L}_{reg}$ w/ sphere prior & 34.9 & 15.8 \\
      $\mathcal{L}_{distr} + \mathcal{L}_{geoc} + \mathcal{L}_{reg}$ w/ sphere prior & 62.7 & 34.6 \\
      $\mathcal{L}_{reg}$ & 26.5 & 14.0 \\ 
      $\mathcal{L}_{distr} +\mathcal{L}_{contr} + \mathcal{L}_{geoc} + \mathcal{L}_{reg}$ & 55.4 & 31.6 \\ \hline
      $\mathcal{L}_{contr} + \mathcal{L}_{geoc} + \mathcal{L}_{reg}$ & 73.5 & 51.6 \\
      $\mathcal{L}_{distr} + \mathcal{L}_{geoc} + \mathcal{L}_{reg}$ & 72.3 & 43.3\\
      \bottomrule
    \end{tabular}
    }
  \end{minipage}
  \hfill
  \begin{minipage}{0.47\linewidth}
    \caption{Keypoint matching scores on horses in PF-Pascal (PFP) and AwAPose (AwA) for  ablations of the model, evaluated using PCK@0.1. 
    The poor performance of the initial prior shows the gains of the optimisation while the sphere prior shows having a more complex prior can improve results. Geometric regularisation, is necessary but not sufficient.}
    \label{table:ablations}
  \end{minipage}
      \vspace{-10pt}   
\end{table}

\begin{table}[h]
  \centering
  \small
   \begin{minipage}{0.65\linewidth}
    \centering
    \resizebox{0.95\textwidth}{!}{
    \begin{tabular}{lrrrrr}
      \toprule
      & Cow & Horse & Sheep & Giraffe & Zebra \\ 
      \midrule
      DINOv2 & 48.7 & 54.7 & 56.3 & 69.9 & 59.0 \\ 
      DINOv2 (pca16) & 46.0 & 53.9 & 48.6 & 67.1 & 59.4\\ \hline
      Ours Contrastive & 48.5 & 41.8 & 43.6 & 40.9 & 36.0\\
      Ours Distributive & 46.4 & 36.6 & 39.9 & 50.5 & 25.4 \\ %
      Ours Contrastive (topk) & 61.5 & 64.3 & 61.3 & 67.4 & 61.3 \\
      Ours Distributive (topk) & 61.4 & 55.7 & 60.4 & 68.0 & 59.1 \\ %
      \bottomrule
    \end{tabular}
    }
  \end{minipage}
    \hfill
  \begin{minipage}{0.32\linewidth}
    \caption{Matching scores for keypoints with left and right equivalents in AwAPose using PCK@0.1. In most instances we improve left-right disambiguation compared to DINOv2.}
    \label{table:symmetry_pck}
    \end{minipage}
    \vspace{-10pt}   
\end{table}

In Table~\ref{table:ablations} we ablate our primary  losses: the geometric regularisations. 
We also test the performance of the initial 3D prior, \ie the representation prior to training, as well as the use of a version of our model that only estimates features using a single sphere (\ie as in \cite{mariotti_improving_2024}), as opposed to a collection of primitives. 
We compare these against both our topk model, as well as the direct model without topk. 
However, it is worth bearing in mind that the  PF-Pascal test set is small, and so to further evaluate these versions of the model, we also ablate them on the AwAPose dataset~\cite{banik_novel_nodate} where a similar trend is seen.
Combining both Contrastive and Distributional approaches worsened performance. Empirically these models appeared to be more susceptible to developing multiple heads and possibly work against one another.

\vspace{-5pt}
\subsection{Limitations}
\vspace{-5pt}
While our approach can approximate the distribution of real image features, there can still be inconsistencies between very similar rendered views. 
For example, the front legs for quadrupeds can frequently change from having front leg features to back leg features to front again. 
There can also be artifacts due to sets of features being very similar, \eg the head of an animal can sometimes gain tail features scattered among the head features. 
This is worse depending on how similar the original backbone features are for that category. 
Finally, our approach is simple by design, but as a result does not explicitly model significant object shape variation or articulations.  
Doing so, would necessitate a significantly more complex model, and our goal is to demonstrate effective feature learning from a simple initial 3D prior.

\vspace{-5pt}
\section{Conclusion}
\vspace{-5pt}
Reliably distinguishing object parts is an open challenge and most previous methods rely on using extensive supervision  to address it. 
We presented a new  lightweight self-supervised approach for extracting more 3D consistent part-centric visual features from a pre-trained feature extractor. 
We achieve this by optimising a category level shape and appearance prior and show that this can be used to train an adapter to disambiguate repeated and symmetric parts for target categories. 
We demonstrate competitive performance on the task of semantic correspondence estimation using two variants of our model on a variety of different object categories.

\clearpage
\newpage 
\bibliographystyle{splncs04}
\bibliography{main}

\appendix
\setcounter{table}{0}
\renewcommand{\thetable}{A\arabic{table}}
\setcounter{figure}{0}
\renewcommand{\thefigure}{A\arabic{figure}}
\section{Additional Results}

Here we present additional results on the AwAPose dataset for five categories: cow, horse, sheep, giraffe, and zebra. In Table~\ref{table:awapose_pck} we see mixed results with our model beating the original DINOv2 features for some categories but not for others, though the results are close.

\begin{table}[h!]
  \centering
  \small
    \centering
        \caption{Keypoint matching scores for five categories in AwAPose including cow, horse, sheep, giraffe and zebra, evaluated using PCK@0.1.}
        \vspace{-5pt}    
    \begin{tabular}{lrrrrr}
      \toprule
      & Cow & Horse & Sheep & Giraffe & Zebra \\ 
      \midrule
      DINOv2 & 60.5 & 65.5 & 61.9 & 76.3 & 69.2 \\
      DINOv2 (pca16) & 56.8 & 64.7 & 57.0 & 73.2 & 68.1\\ \hline
      Ours Contrastive & 49.6 & 51.6 & 44.5 & 46.8 & 46.2 \\
      Ours Distributive & 48.7 & 43.3 & 40.3 & 56.5 & 27.0 \\ %
      Ours Contrastive (topk) & 64.2 & 68.5 & 62.2 & 71.9 &  68.9 \\
      Ours Distributive (topk) & 63.8 & 63.5 & 61.5 & 72.9 & 66.7 \\ %
      \bottomrule
    \end{tabular}
    \vspace{-5pt}    

    \label{table:awapose_pck}
\end{table}

\begin{figure}[h]
    \centering

    \begin{minipage}{0.9\textwidth}
        \centering
        \includegraphics[width=0.24\linewidth]{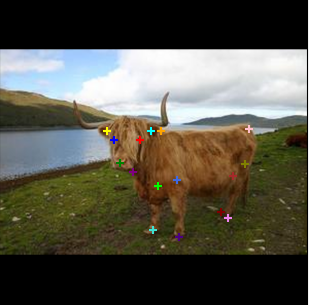}
        \includegraphics[width=0.24\linewidth]{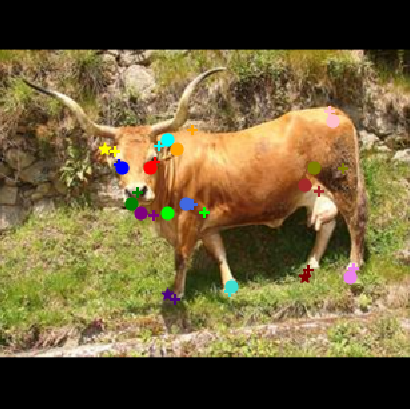}
        \includegraphics[width=0.24\linewidth]{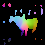}
        \includegraphics[width=0.24\linewidth]{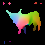}
        \\
        \includegraphics[width=0.24\linewidth]{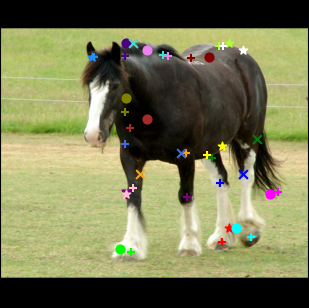}
        \includegraphics[width=0.24\linewidth]{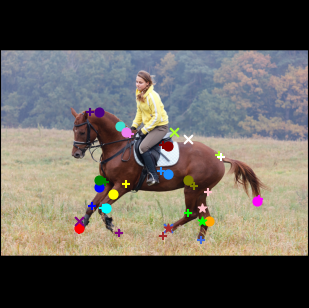}
        \includegraphics[width=0.24\linewidth]{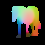}
        \includegraphics[width=0.24\linewidth]{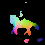}
        \\
        \includegraphics[width=0.24\linewidth]{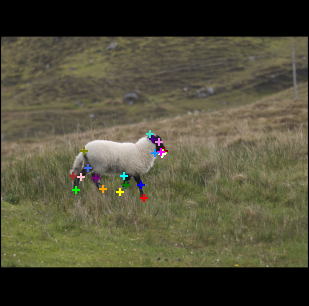}
        \includegraphics[width=0.24\linewidth]{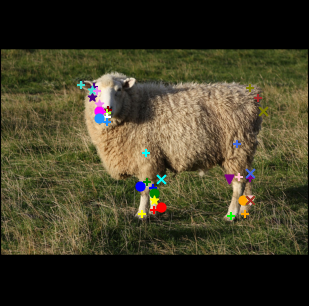}
        \includegraphics[width=0.24\linewidth]{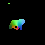}
        \includegraphics[width=0.24\linewidth]{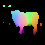}
        \\
        \includegraphics[width=0.24\linewidth]{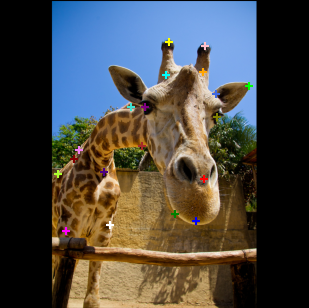}
        \includegraphics[width=0.24\linewidth]{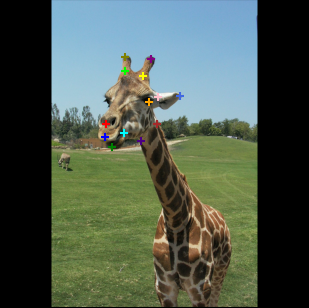}
        \includegraphics[width=0.24\linewidth]{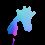}
        \includegraphics[width=0.24\linewidth]{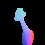}
        \\
        \includegraphics[width=0.24\linewidth]{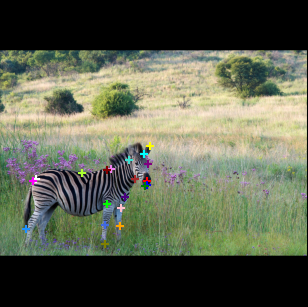}
        \includegraphics[width=0.24\linewidth]{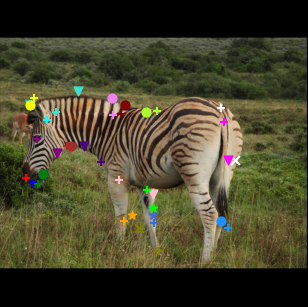}
        \includegraphics[width=0.24\linewidth]{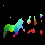}
        \includegraphics[width=0.24\linewidth]{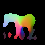}
    \end{minipage}
      \vspace{4pt}
        \caption{Results from the PFPascal and AwAPose test sets are presented in the first two columns and their corresponding adapted features in the second two columns. '+' indicates the labelled keypoint, while predictions are indicated by a dot where both DINOv2 and our model were successful, a downward arrow where DINOv2 succeeded but ours failed, a star where ours succeeded and DINOv2 failed, and a 'X' where both models failed. Looking at the first row we see our model has successfully distinguished the four legs of the cow and this has enabled it to differentiate the two front legs. It also now succeeds on the back leg, made difficult due to the ambiguity of the original keypoint. It has also successfully distinguished the two eyes.}
        \vspace{-5pt}
    \label{fig:pfpascal_example}
\end{figure}

Figure~\ref{fig:pfpascal_example} shows the keypoint transfer results for PFPascal and AwAPose examples across five categories. We see the adapted features are able to use the learned 3D position predictions to improve keypoint assignment better disambiguating left from right than the original DINOv2 features.

In Table~\ref{table:pfascal_results_more_categories} we show PFPascal results across a diverse set of categories. The Distributive model's direct features are more robust on average than the Contrastive model's and Distributive performs 4\% better using topk.

\begin{table}[h]
  \centering
  \small
    \centering
    \caption{
      Evaluation on PF-Pascal for additional categories, where for each category we calculate
      PCK@0.1 for cosine similarity nearest neighbour from source to
      target. We follow other papers and use the image resolution when
      calculating the threshold rather than the bounding box. For the topk approach we used k=5 matches.
    }
    \resizebox{1.0\textwidth}{!}{
    \begin{tabular}{lcccccccccccc}
      \toprule
      & \faIcon{bicycle} & \faIcon{wine-bottle} & \faIcon{crow} & \faIcon{car} & \faIcon{cat} & \Cow & \faIcon{dog} & \faIcon{horse} & \faIcon{walking} & \Plant &\Sheep & Avg\\ 
      \midrule
      DINOv2~\cite{oquab_dinov2_2023} & 86.0 & 56.3 & 85.9 &  90.4 & 73.4 & 87.5 & 72.4 & 84.3 & 62.1 & 51.2 & 80.0 & 75.4 \\
      DINOv2~\cite{oquab_dinov2_2023} PCA16 & 85.4 & 48.4 & 85.9 & 88.0 & 64.1 & 83.3 & 60.1 & 85.5 & 54.8 & 36.6 & 80.0  & 70.2 \\ \hline
      SphericalMaps~\cite{mariotti_improving_2024} (sphere only) & 66.6 & 21.9 & 52.6 & 58.7 & 79.9 & 64.6 & 68.7 & 61.5 & 29.4 & 19.5 & 80.0 & 54.9 \\
      SphericalMaps~\cite{mariotti_improving_2024} & 90.6 & 78.1 & 87.2 & 94.6 & 84.2 & 91.7 & 84.7 & 85.5 & 71.8 & 65.9 & 80.0 & 83.1 \\
 \hline
      Ours Contrastive & 51.7 & 17.2 & 35.9 & 59.3 & 41.3 & 81.6 & 52.8 & 73.5 & 52.0 & 22.0 & 60.0 & 43.9 \\
      Ours Distributive & 48.9 & 18.8 & 55.1 & 85.3 & 73.9 & 83.3 & 44.2 & 72.3 & 78.5 & 46.3 & 60.0 & 60.6 \\
      Ours Contrastive (with topk) & 86.0 & 56.3 & 79.5 & 87.4 & 64.7 & 97.9 & 76.1 & 80.7 & 60.5 & 36.6 & 80.0 & 73.2\\
      Ours Distributive (with topk) & 86.6 & 46.9 & 82.0 & 86.8 & 82.6 & 93.8 & 73.0 & 84.3 & 63.8 & 56.1 & 80.0 & 76.4 \\
      \bottomrule
    \end{tabular}
    }
  \hfill
    \label{table:pfascal_results_more_categories}
\end{table}

\vspace{-5pt}
\section{Implementation Details}
\vspace{-5pt}

\subsection{Learning Shape and Texture}
\noindent\textbf{Training.} In addition to the topology, we assume an approximate distribution of camera views reflecting our knowledge of the training dataset. 
For quadrupeds and other large animals this will typically be either a uniform or normal distribution for rotation about the vertical y-axis, while for household objects like cups or small animals like insects we might assume a uniform distribution across all axes. A normal distribution would be more appropriate for datasets where the images are biased toward a particular view, \eg frontal for cats.
To avoid our model collapsing to a degenerate solution we enforce global scale invariance by rescaling all renders to fill the image. 
For the contrastive loss we treat each symmetric half of the mesh as separate views. 
This avoids the possible contradiction of saying that of two equivalent symmetric vertices, the features on one should be more similar to an anchor point than those of the other. To fully mimic real images, synthetic images are rendered on to random real backgrounds before passing to the feature extractor.

To ensure every sample can be transported using optimal transport and is not limited by the shapes or occlusions present in the images of the current minibatch, and further to better approximate the distribution for the whole dataset, we fit a Gaussian mixture model (GMM), prior to training, to the feature distribution of real images. We then sample the mixture model to match the number of samples to those of the synthetic features.
We find the GMM fits the feature distribution well for when dimensionality is reduced to 16 dimensions with PCA.

The feature reconstruction uses an Adam optimiser with a learning rate of 0.05, scheduled with $\gamma$=0.1 every 40 steps, and trained up to 200 iterations where each iteration uses a batch of 12 views. The hyperparameters used were $\lambda_{\sigma}=1e-2, \lambda_{\delta}=1e-2, \lambda_{\zeta}=1.0, \lambda_{coll}=1.0$, and contrastive constant $\alpha=0.2$ with contrastive radius 2.
For our adapter we use a U-NET with an initial lr of 5e-4, reducing every 1000 steps with $\gamma=0.5$, and confidence weight $\lambda_c=1.0$. To help the adapter generalise we augment the generated synthetic feature maps using translations, crops, rotations, distortions, shears and patch and border masking.
Input images were 630$\times$630 and we used DINOv2 as our feature extractor $\phi()$. Each ellipsoid consisted of 2562 vertices.
For optimal transport we use POT~\cite{flamary_pot_2021} with cosine distance and regularisation $\epsilon=1e-3$. 

Due to the random initialisation of the model, results can vary. 
As we do not require zero-shot accuracy to build a canonical template, we run the template learning ten times and choose the model that minimises the loss. We use early stopping so as not to waste time on bad runs, stopping early if the moving average of 10 iterations does not drop below the previous minimum average.
Each run therefore takes a variable amount of time from one to three hours on an A6000 GPU, depending also on the number of ellipsoids. For the horse category for example, the distributive model took 2 hours and 24 minutes while one of the shorter runs took 1 hour and 8 minutes. For the car category, which used a single ellipsoid, one of the longer distributive runs took 57 minutes.
In line with previous work we evaluate PCK@0.1 using the image resolution as the bounding box for PF-Pascal when comparing to~\cite{mariotti_improving_2024}.

\noindent\textbf{Datasets.} To train the canonical representations we use the 3DFauna dataset~\cite{li_learning_2024} for horse, sheep, and cow categories, and we use Freiburg for cars \cite{sedaghat_unsupervised_2015}. For the bird category we use CUB~\cite{wah2011caltech} while for spiders we use images downloaded from iNaturalist \cite{noauthor_inaturalist_nodate}. Giraffe was trained on Apt-36k~\cite{apt36k} and zebra used the LASSIE zebra training set~\cite{yao_lassie_2022}. For cats we used a combination of Animal3D, BANMo, Apt-36k, and Oxford Cats~\cite{xu2023animal3d,yang2022banmo,apt36k, catsdogs}, for dogs we used Animal3d and APT~\cite{xu2023animal3d,apt36k}, for person we used the Leeds Sports Pose dataset~\cite{Johnson11}, while for the other categories we used the train set of SPair-17k~\cite{min_spair-71k_nodate}.
We obtained segmentation masks at training time from SAM~\cite{kirillov2023segment} and bounding boxes from GroundingDINO~\cite{liu2024grounding}. 
For spiders and potted plant, the masks were sufficiently poor and were interactively masked. As we want the category to be as centered as possible in the image we removed close-up images of gears when training the bicycle category on SPair-71k images~\cite{min_spair-71k_nodate}. The displayed spider results were obtained without perturbation of the skeleton.

\subsection{Adapter Learning}
We adopt for the adapter architecture the U-Net used in~\cite{song2021score} as implemented by~\cite{karras2022elucidating} with latent dimension 128 and 2 blocks. The inputs are the 2D maps of features we want to adapt, their dimension reduced using PCA as for training the template model. We generate our training data by rendering the template model to 180 random views with random poses and articulations. We extract features for each of these renders and render vertex positions to produce GT 3D position data, as well as masks. We transport the extracted features to real features using optimal transport, the real features being sampled from the trained GMM. For our results we used a view prior biased toward the front.

\end{document}